# On the Equivalence of Holographic and Complex Embeddings for Link Prediction[*]

Katsuhiko Hayashi
NTT Communication Science Laboratories, Seika-cho, Kyoto 619 0237, Japan
hayashi.katsuhiko@lab.ntt.co.jp

Masashi Shimbo
Nara Institute of Science and Technology, Ikoma, Nara 630 0192, Japan
shimbo@is.naist.jp

**Abstract**

We show the equivalence of two state-of-the-art models for link prediction/knowledge graph completion: Nickel et al's holographic embeddings and Trouillon et al.'s complex embeddings. We first consider a spectral version of the holographic embeddings, exploiting the frequency domain in the Fourier transform for efficient computation. The analysis of the resulting model reveals that it can be viewed as an instance of the complex embeddings with a certain constraint imposed on the initial vectors upon training. Conversely, any set of complex embeddings can be converted to an equivalent set of holographic embeddings, in the sense that they give the same scores for any relation triples up to scaling.

## 1 Introduction

Recently, there have been efforts to build and maintain large-scale knowledge bases represented in the form of a graph (*knowledge graph*) (Auer et al., 2007; Bollacker et al., 2008; Suchanek et al., 2007). Although these knowledge graphs contain billions of relational facts, they are known to be incomplete. *Knowledge graph completion* (KGC) (Nickel et al., 2015) aims at augmenting missing knowledge in an incomplete knowledge graph automatically. It can be viewed as a task of *link prediction* (Liben-Nowell and Kleinberg, 2003; Hasan and Zaki, 2011) studied in the field of statistical relational learning (Getoor and Taskar, 2007). In recent years, methods based on vector embeddings of graphs have been actively pursued as a scalable approach to KGC (Bordes et al., 2011; Socher et al., 2013; Guu et al., 2015; Yang et al., 2015; Nickel et al., 2016; Trouillon et al., 2016b).

In this paper, we investigate the connection between two models of graph embeddings that have emerged along this line of research: The *holographic embeddings* (Nickel et al., 2016) and the *complex embeddings* (Trouillon et al., 2016b). These models are simple yet achieve the current state-of-the-art performance in KGC.

---

[*]This is a slightly modified version of the paper that appeared in ACL 2017 (Hayashi and Shimbo, 2017).

We begin by showing that holographic embeddings can be trained entirely in the frequency domain induced by the Fourier transform, thereby reducing the time needed to compute the scoring function from $O(n \log n)$ to $O(n)$, where $n$ is the dimension of the embeddings.

The analysis of the resulting training method reveals that the Fourier transform of holographic embeddings can be regarded as an instance of the complex embeddings, with a specific constraint (viz. conjugate symmetry property) cast on the initial values.

Conversely, it is shown that every set of complex embeddings has a set of holographic embeddings (with real vectors) that is equivalent, in the sense that their scoring functions are equal up to scaling.

## 2 Preliminaries

Let $i$ denote the imaginary unit, $\mathbb{R}$ be the set of real values, and $\mathbb{C}$ the set of complex values. We write $[\mathbf{v}]_j$ to denote the $j$th component of vector $\mathbf{v}$. A superscript T (e.g., $\mathbf{v}^\mathrm{T}$) represents vector/matrix transpose. For a complex scalar $z$, vector $\mathbf{z}$, and matrix $\mathbf{Z}$, $\bar{z}$, $\bar{\mathbf{z}}$, and $\bar{\mathbf{Z}}$ represent their complex conjugate, and $\mathrm{Re}(z)$, $\mathrm{Re}(\mathbf{z})$, and $\mathrm{Re}(\mathbf{Z})$ denote their real parts, respectively.

Let $\mathbf{x} = [x_0 \cdots x_{n-1}]^\mathrm{T} \in \mathbb{R}^n$ and $\mathbf{y} = [y_0 \cdots y_{n-1}]^\mathrm{T} \in \mathbb{R}^n$. Note that the vector indices start from 0 for notational convenience. The *circular convolution* of $\mathbf{x}$ and $\mathbf{y}$, denoted by $\mathbf{x} * \mathbf{y}$, is defined by

$$[\mathbf{x} * \mathbf{y}]_j = \sum_{k=0}^{n-1} x_{[(j-k) \bmod n]} y_k, \tag{1}$$

where mod denotes modulus. Likewise, *circular correlation* $\mathbf{x} \star \mathbf{y}$ is defined by

$$[\mathbf{x} \star \mathbf{y}]_j = \sum_{k=0}^{n-1} x_{[(k-j) \bmod n]} y_k. \tag{2}$$

While circular convolution is commutative, circular correlation is not; i.e., $\mathbf{x} * \mathbf{y} = \mathbf{y} * \mathbf{x}$, but $\mathbf{x} \star \mathbf{y} \neq \mathbf{y} \star \mathbf{x}$ in general. As it can be verified with Eqs. (1) and (2), $\mathbf{x} \star \mathbf{y} = \mathrm{flip}(\mathbf{x}) * \mathbf{y}$, where $\mathrm{flip}(\mathbf{x}) = [x_{n-1} \cdots x_0]^\mathrm{T}$ is a vector obtained by arranging the components of $\mathbf{x}$ in reverse.

For $n$-dimensional vectors, naively computing circular convolution/correlation by Eqs. (1) and (2) requires $O(n^2)$ multiplications. However, we can take advantage of the Fast Fourier Transform (FFT) algorithm to accelerate the computation: For circular convolution, first compute the discrete Fourier transform (DFT) of $\mathbf{x}$ and $\mathbf{y}$, and then compute the inverse DFT of their elementwise vector product, i.e.,

$$\mathbf{x} * \mathbf{y} = \mathfrak{F}^{-1}(\mathfrak{F}(\mathbf{x}) \odot \mathfrak{F}(\mathbf{y})),$$

where $\mathfrak{F} : \mathbb{R}^n \to \mathbb{C}^n$ and $\mathfrak{F}^{-1} : \mathbb{C}^n \to \mathbb{R}^n$ respectively denote the DFT and inverse DFT, and $\odot$ denotes the elementwise product. DFT and inverse DFT can be computed in $O(n \log n)$ time with the FFT algorithm, and thus the computation time for circular convolution is also $O(n \log n)$. The same can be said of circular correlation. Since $\mathfrak{F}(\mathrm{flip}(\mathbf{x})) = \overline{\mathfrak{F}(\mathbf{x})}$, we have

$$\mathbf{x} \star \mathbf{y} = \mathfrak{F}^{-1}(\overline{\mathfrak{F}(\mathbf{x})} \odot \mathfrak{F}(\mathbf{y})). \tag{3}$$

By analogy to how the Fourier transform is used in signal processing, the original real space $\mathbb{R}^n$ is called the "time" domain, and the complex space $\mathbb{C}^n$ where DFT vectors reside is called the "frequency" domain.

## 3 Holographic embeddings for knowledge graph completion

### 3.1 Knowledge graph completion

Let $\mathcal{E}$ and $\mathcal{R}$ be the finite sets of *entities* and *(binary) relations* over entities, respectively. For each relation $r \in \mathcal{R}$ and each pair $s, o \in \mathcal{E}$ of entities, we are interested in whether $r(s, o)$ holds[1] or not; we write $r(s, o) = +1$ if it holds, and $r(s, o) = -1$ if not. To be precise, given a *training set* $\mathcal{D} \subset \mathcal{R} \times \mathcal{E} \times \mathcal{E} \times \{-1, +1\}$ such that $(r, s, o, y) \in \mathcal{D}$ indicates $y = r(s, o)$, our task is to design a *scoring function* $f : \mathcal{R} \times \mathcal{E} \times \mathcal{E} \to \mathbb{R}$ such that for each of the triples $(r, s, o)$ not observed in $\mathcal{D}$, function $f$ should give a higher value if $r(s, o) = +1$ is more likely, and a smaller value for those that are less likely. If necessary, $f(r, s, o)$ can be converted to probability by $P[r(s, o) = +1] = \sigma(f(r, s, o))$, where $\sigma : \mathbb{R} \to (0, 1)$ is a sigmoid function.

Dataset $\mathcal{D}$ can be regarded as a directed graph in which nodes represent entities $\mathcal{E}$ and edges are labeled by relations $\mathcal{R}$. Thus, the task is essentially that of *link prediction* (Liben-Nowell and Kleinberg, 2003; Hasan and Zaki, 2011). Often, it is also called *knowledge graph completion*.

### 3.2 Holographic embeddings (HolE)

Nickel et al. (2016) proposed *holographic embeddings* (HolE) for knowledge graph completion. Using training data $\mathcal{D}$, this method learns the vector embeddings $\mathbf{e}_k \in \mathbb{R}^n$ of entities $k \in \mathcal{E}$ and the embeddings $\mathbf{w}_r \in \mathbb{R}^n$ of relations $r \in \mathcal{R}$. The score for triple $(r, s, o)$ is then given by

$$f_{\text{HolE}}(r, s, o) = \mathbf{w}_r \cdot (\mathbf{e}_s \star \mathbf{e}_o). \tag{4}$$

Eq. (4) can be evaluated in time $O(n \log n)$ if $\mathbf{e}_s \star \mathbf{e}_o$ is computed by Eq. (3).

## 4 Spectral training of HolE

To compute the circular correlation in the scoring function of Eq. (4) efficiently, Nickel et al. (2016) used Eq. (3) in Section 2 and FFT. In this section, we extend this technique further, and consider training HolE solely in the frequency domain. That is, real-valued embeddings $\mathbf{e}_k, \mathbf{w}_r \in \mathbb{R}^n$ in the original "time" domain are abolished, and instead we train their DFT counterparts $\boldsymbol{\varepsilon}_k = \mathfrak{F}(\mathbf{e}_k) \in \mathbb{C}^n$ and $\boldsymbol{\omega}_k = \mathfrak{F}(\mathbf{w}_r) \in \mathbb{C}^n$ in the frequency domain. This formulation eliminates the need of FFT and inverse FFT, which are the major computational bottleneck in HolE. As a result, Eq. (4) can be computed in time $O(n)$ directly from $\boldsymbol{\varepsilon}_k$ and $\boldsymbol{\omega}_k$.

Indeed, equivalent counterparts in the frequency domain exist for not only convolution/correlation but all other computations needed for HolE: scalar multiplication, summation (needed when vectors are updated by stochastic gradient descent), and dot product (used in Eq. (4)). The frequency-domain equivalents for these operations are summarized in Table 1. All of these can be performed efficiently (in linear time) in the frequency domain.

---

[1] Depending on the context, letter $r$ is used either as an index to an element in $\mathcal{R}$ or the binary relation it signifies.

Table 1: Correspondence between operations in time and frequency domains. $\mathbf{r} \leftrightarrow \boldsymbol{\rho}$ indicates $\boldsymbol{\rho} = \mathfrak{F}(\mathbf{r})$ (and also $\mathbf{r} = \mathfrak{F}^{-1}(\boldsymbol{\rho})$).

| operation | time | | frequency |
|---|---|---|---|
| scalar mult. | $\alpha \mathbf{x}$ | $\longleftrightarrow$ | $\alpha \mathfrak{F}(\mathbf{x})$ |
| summation | $\mathbf{x} + \mathbf{y}$ | $\longleftrightarrow$ | $\mathfrak{F}(\mathbf{x}) + \mathfrak{F}(\mathbf{y})$ |
| flip | $\mathrm{flip}(\mathbf{x})$ | $\longleftrightarrow$ | $\overline{\mathfrak{F}(\mathbf{x})}$ |
| convolution | $\mathbf{x} * \mathbf{y}$ | $\longleftrightarrow$ | $\mathfrak{F}(\mathbf{x}) \odot \mathfrak{F}(\mathbf{y})$ |
| correlation | $\mathbf{x} \star \mathbf{y}$ | $\longleftrightarrow$ | $\overline{\mathfrak{F}(\mathbf{x})} \odot \mathfrak{F}(\mathbf{y})$ |
| dot product | $\mathbf{x} \cdot \mathbf{y}$ | $=$ | $\frac{1}{n} \mathfrak{F}(\mathbf{x}) \cdot \mathfrak{F}(\mathbf{y})$ |

In particular, the following relation holds for the dot product between any "time" vectors $\mathbf{x}, \mathbf{y} \in \mathbb{R}^n$.

$$\mathbf{x} \cdot \mathbf{y} = \frac{1}{n} \mathfrak{F}(\mathbf{x}) \cdot \mathfrak{F}(\mathbf{y}), \tag{5}$$

where the dot product on the right-hand side is the complex inner product defined by $\mathbf{a} \cdot \mathbf{b} = \overline{\mathbf{a}}^\mathrm{T} \mathbf{b}$. Eq. (5) is known as Parseval's theorem (also called the *power theorem* in (Smith, 2007)), and it states that dot products in two domains are equal up to scaling.

After embeddings $\boldsymbol{\varepsilon}_k, \boldsymbol{\omega}_r \in \mathbb{C}^n$ are learned in the frequency domain, their time-domain counterparts $\mathbf{e}_k = \mathfrak{F}^{-1}(\boldsymbol{\varepsilon}_k)$ and $\mathbf{w}_r = \mathfrak{F}^{-1}(\boldsymbol{\omega}_r)$ can be recovered if needed, but this is not required as far as computation of the scoring function is concerned. Thanks to Parseval's theorem, Eq. (4) can be directly computed from the frequency vectors $\boldsymbol{\varepsilon}_k, \boldsymbol{\omega}_r \in \mathbb{C}^n$ by

$$f_{\mathrm{HolE}}(r, s, o) = \frac{1}{n} \boldsymbol{\omega}_r \cdot (\overline{\boldsymbol{\varepsilon}_s} \odot \boldsymbol{\varepsilon}_o). \tag{6}$$

### 4.1 Conjugate symmetry of spectral components

A complex vector $\boldsymbol{\xi} = [\xi_0 \cdots \xi_{n-1}]^\mathrm{T} \in \mathbb{C}^n$ is said to be *conjugate symmetric* (or *Hermitian*) if $\xi_j = \overline{\xi_{[(n-j) \bmod n]}}$ for $j = 0, \ldots, n-1$, or, in other words, if it can be written in the form

$$\boldsymbol{\xi} = \begin{cases} \begin{bmatrix} \xi_0 & \boldsymbol{\gamma} & \mathrm{flip}(\overline{\boldsymbol{\gamma}}) \end{bmatrix}^\mathrm{T}, & \text{if } n \text{ is odd,} \\ \begin{bmatrix} \xi_0 & \boldsymbol{\gamma} & \xi_{n/2} & \mathrm{flip}(\overline{\boldsymbol{\gamma}}) \end{bmatrix}^\mathrm{T}, & \text{if } n \text{ is even,} \end{cases}$$

for some $\boldsymbol{\gamma} \in \mathbb{C}^{\lceil n/2 \rceil - 1}$ and $\xi_0, \xi_{n/2} \in \mathbb{R}$.

The DFT $\mathfrak{F}(\mathbf{x})$ is conjugate symmetric if and only if $\mathbf{x}$ is a real vector. Thus, maintaining conjugate symmetry of "frequency" vectors is the key to ensure their "time" counterparts remain in real space. Below, we verify that this property is indeed preserved with stochastic gradient descent. Moreover, conjugate symmetry provides a sufficient condition under which dot product takes a real value. It also has implications on space requirement. These topics are also covered in the rest of this section.

## 4.2 Vector initialization and update in frequency domain

Typically, at the beginning of training HolE, each individual embedding is initialized by a random vector. When we train HolE in the frequency domain, we could first generate a random real vector, regard them as a HolE vector in the time domain, and compute its DFT as the initial value in the frequency domain. An alternative, easier approach is to directly generate a random complex vector that is conjugate symmetric, and use it as the initial frequency vector. This guarantees the inverse DFT to be a real vector, i.e., there exists a valid corresponding image in the time domain.

Given a training set $\mathcal{D}$ (see Section 3.1), HolE is trained by minimizing the following objective function over parameter matrix $\mathbf{\Theta} = [\mathbf{e}_1 \cdots \mathbf{e}_{|\mathcal{E}|} \, \mathbf{w}_1 \cdots \mathbf{w}_{|\mathcal{R}|}] \in \mathbb{R}^{n \times (|\mathcal{E}|+|\mathcal{R}|)}$:

$$\sum_{(r,s,o,y) \in \mathcal{D}} \log\{1 + \exp(-y f_{\text{HolE}}(r,s,o))\} + \lambda \|\mathbf{\Theta}\|_F^2 \tag{7}$$

where $\lambda \in \mathbb{R}$ is the hyperparameter controlling the degree of regularization, and $\|\cdot\|_F$ denotes the Frobenius norm.

In our version of spectral training of HolE, the parameters matrix consists of frequency vectors $\mathbf{\varepsilon}_k$ and $\mathbf{\omega}_r$ instead of $\mathbf{e}_k$ and $\mathbf{w}_r$, i.e., $\mathbf{\Theta} = [\mathbf{\varepsilon}_1 \cdots \mathbf{\varepsilon}_{|\mathcal{E}|} \, \mathbf{\omega}_1 \cdots \mathbf{\omega}_{|\mathcal{R}|}] \in \mathbb{C}^{n \times (|\mathcal{E}|+|\mathcal{R}|)}$. Let us discuss the stochastic gradient descent (SGD) update with respect to these frequency vectors. In particular, we are interested in whether conjugate symmetry of vectors is preserved by the update.

Suppose vectors $\mathbf{\omega}_r, \mathbf{\varepsilon}_s, \mathbf{\varepsilon}_o$ are conjugate symmetric. Neglecting the contribution from the regularization term[2] in Eq. (7), we see that in an SGD update step, $\alpha \partial f_{\text{HolE}}/\partial \mathbf{\omega}_r$, $\alpha \partial f_{\text{HolE}}/\partial \mathbf{\varepsilon}_s$, and $\alpha \partial f_{\text{HolE}}/\partial \mathbf{\varepsilon}_o$ are respectively subtracted from $\mathbf{\omega}_r, \mathbf{\varepsilon}_s, \mathbf{\varepsilon}_o$, where $\alpha \in \mathbb{R}$ is a scalar. Noting the equalities

$$\mathbf{w}_r \cdot (\mathbf{e}_s \star \mathbf{e}_o) = \mathbf{e}_s \cdot (\mathbf{w}_r \star \mathbf{e}_o) = \mathbf{e}_o \cdot (\mathbf{w}_r * \mathbf{e}_s)$$

(see (Nickel et al., 2016, Eq. (12), p. 1958)) and their frequency counterparts

$$\mathbf{\omega}_r \cdot (\overline{\mathbf{\varepsilon}_s} \odot \mathbf{\varepsilon}_o) = \mathbf{\varepsilon}_s \cdot (\overline{\mathbf{\omega}_r} \odot \mathbf{\varepsilon}_o) = \mathbf{\varepsilon}_o \cdot (\mathbf{\omega}_r \odot \mathbf{\varepsilon}_s),$$

obtained through the translation of Table 1, we can derive

$$\frac{\partial f_{\text{HolE}}}{\partial \mathbf{\omega}_r} = \overline{\mathbf{\varepsilon}_s} \odot \mathbf{\varepsilon}_o,$$

$$\frac{\partial f_{\text{HolE}}}{\partial \mathbf{\varepsilon}_s} = \overline{\mathbf{\omega}_r} \odot \mathbf{\varepsilon}_o,$$

$$\frac{\partial f_{\text{HolE}}}{\partial \mathbf{\varepsilon}_o} = \mathbf{\omega}_r \odot \mathbf{\varepsilon}_s.$$

Thus, conjugation and elementwise product are used in an SGD update, as well as scalar multiplication and summation. And it is straightforward to verify that all these operations preserve conjugate symmetry. It follows that if $\mathbf{\omega}_r, \mathbf{\varepsilon}_s, \mathbf{\varepsilon}_o$ are initially conjugate symmetric, they will remain so during the course of training, which assures that the inverse DFTs of the learned embeddings are real vectors.

---

[2] It can be easily verified that the contribution from the regularization term to SGD update does not violate conjugate symmetry.

### 4.3 Real-valued dot product

In the scoring function of HolE (Eq. (4)), dot product is used for generating a real-valued "score" out of two vectors, $\mathbf{w}_r$ and $\mathbf{e}_s \odot \mathbf{e}_o$. Likewise, in Eq. (6), the dot product is applied to $\boldsymbol{\omega}_r$ and $\boldsymbol{\varepsilon}_s \odot \boldsymbol{\varepsilon}_o$, which are complex-valued. However, provided that the conjugate symmetry of these vectors is maintained, their dot product is always real. This follows from Parseval's theorem; the inverse DFTs of these frequency vectors are real, and thus their dot product is also real. Therefore, the dot product of the corresponding frequency vectors is real as well, according to Eq. (5).

### 4.4 Space requirement

A general complex vector $\boldsymbol{\xi} \in \mathbb{C}^n$ can be stored in memory as $2n$ floating-point numbers, i.e., one each for the real and imaginary part of a component. In our spectral representation of HolE, however, the first $\lfloor n/2 \rfloor$ components suffice to specify the frequency vector $\boldsymbol{\xi}$, since the vector is conjugate symmetric. Moreover, $\xi_0$ (and $\xi_{n/2}$ if $n$ is even) are real values. Thus, a spectral representation of HolE can be specified with exactly $n$ floating-point numbers, which can be stored in the same amount of memory as needed by the original HolE.

## 5 Relation to Trouillon et al.'s complex embeddings

### 5.1 Complex embeddings (ComplEx)

Trouillon et al. (2016b) proposed a model of embedding-based knowledge graph completion, called *complex embeddings* (ComplEx). The objective is similar to Nickel et al.'s; the embeddings $\mathbf{e}_k$ of entities and $\mathbf{w}_r$ of relations are to be learned. In their model, however, these vectors are complex-valued, and are based on the eigendecomposition of complex matrix $\mathbf{X}_r = \mathbf{E}\mathbf{W}_r\overline{\mathbf{E}}^{\mathrm{T}}$ that encodes relation $r \in \mathcal{R}$ over pairs of entities, where $\mathbf{X}_r \in \mathbb{C}^{|\mathcal{E}|\times|\mathcal{E}|}$, $\mathbf{E} = [\mathbf{e}_1,\ldots,\mathbf{e}_{|\mathcal{E}|}]^{\mathrm{T}} \in \mathbb{C}^{|\mathcal{E}|\times n}$, and $\mathbf{W}_r = \mathrm{diag}(\mathbf{w}_r) \in \mathbb{C}^{n\times n}$ is a diagonal matrix (with diagonal elements $\mathbf{w}_r \in \mathbb{C}^n$). In practice, $\mathbf{X}_r$ needs to be a real matrix, because its $(r,s)$-component must define the score for $r(s,o)$. To this end, Trouillon et al. extracted the real part, i.e., $\mathbf{X}_r = \mathrm{Re}(\mathbf{E}\mathbf{W}_r\overline{\mathbf{E}}^{\mathrm{T}})$. This approach can be advocated by the fact that any real matrix $\mathbf{X}_r$ can be expressed in this form (Trouillon et al., 2016a).

With this formulation, the score for triple $(r,s,o)$ is thus given by

$$f_{\mathrm{ComplEx}}(r,s,o) = \mathrm{Re}\left(\sum_{j=0}^{n-1} [\mathbf{w}_r]_j [\mathbf{e}_s]_j \overline{[\mathbf{e}_o]_j}\right). \tag{8}$$

### 5.2 Equivalence of holographic and complex embeddings

Now let us rewrite Eq. (8). Noting the definition of complex dot product, i.e., $\mathbf{a}\cdot\mathbf{b} = \overline{\mathbf{a}}^{\mathrm{T}}\mathbf{b}$, we have

$$\sum_{j=0}^{n-1} [\mathbf{w}_r]_j [\mathbf{e}_s]_j \overline{[\mathbf{e}_o]_j} = (\mathbf{e}_s \odot \overline{\mathbf{e}_o})^{\mathrm{T}}\mathbf{w}_r$$

$$= \overline{(\mathbf{e}_s \odot \overline{\mathbf{e}_o})} \cdot \mathbf{w}_r \qquad (\because \mathbf{a}\cdot\mathbf{b} = \bar{\mathbf{a}}^\mathrm{T}\mathbf{b})$$

$$= \overline{(\overline{\mathbf{e}_s} \odot \mathbf{e}_o)} \cdot \mathbf{w}_r$$

$$= \overline{\mathbf{w}_r \cdot (\overline{\mathbf{e}_s} \odot \mathbf{e}_o)} \qquad (\because \mathbf{a}\cdot\mathbf{b} = \overline{\mathbf{b}\cdot\mathbf{a}})$$

and since $\mathrm{Re}(\mathbf{z}) = \mathrm{Re}(\bar{\mathbf{z}})$,

$$\mathrm{Re}(\overline{\mathbf{w}_r \cdot (\overline{\mathbf{e}_s} \odot \mathbf{e}_o)}) = \mathrm{Re}(\mathbf{w}_r \cdot (\overline{\mathbf{e}_s} \odot \mathbf{e}_o)).$$

Thus, Eq. (8) can be written as

$$f_{\mathrm{ComplEx}}(r, s, o) = \mathrm{Re}\left(\mathbf{w}_r \cdot (\overline{\mathbf{e}_s} \odot \mathbf{e}_o)\right). \tag{9}$$

Here, a marked similarity is noticeable between Eq. (9) and Eq. (6), the scoring function of our spectral version of HolE (spectral HolE); ComplEx extracts the real part of complex dot product, whereas in the spectral HolE, dot product is guaranteed to be real because all embeddings satisfy conjugate symmetry. Indeed, Eq. (6) can be equally written as

$$f_{\mathrm{HolE}}(r, s, o) = \frac{1}{n} \mathrm{Re}\left(\boldsymbol{\omega}_r \cdot (\overline{\boldsymbol{\varepsilon}_s} \odot \boldsymbol{\varepsilon}_o)\right), \tag{10}$$

although the operator $\mathrm{Re}(\cdot)$ in this formula is redundant.

Conversely, given a set of complex embeddings for entities and relations, we can construct their equivalent holographic embeddings, in the sense that $f_{\mathrm{ComplEx}}(r, s, o) = c f_{\mathrm{HolE}}(r, s, o)$ for every $r, s, o$, with a constant $c > 0$. For each $n$-dimensional complex embeddings $\mathbf{x} \in \{\mathbf{e}_k\}_{k \in \mathcal{E}} \cup \{\mathbf{w}_r\}_{r \in \mathcal{R}} \subset \mathbb{C}^n$ computed by ComplEx, we make a corresponding HolE $\mathfrak{h}(\mathbf{x}) \in \mathbb{R}^{2n+1}$ as follows: For a given complex embedding $\mathbf{x} = [x_0 \cdots x_{n-1}] \in \mathbb{C}^n$, first compute $\mathfrak{s}(\mathbf{x}) \in \mathbb{C}^{2n+1}$ by

$$\begin{aligned}\mathfrak{s}(\mathbf{x}) &= \begin{bmatrix} 0 & x_0 & \cdots & x_{n-1} & \overline{x_{n-1}} & \cdots & \overline{x_0}\end{bmatrix}^\mathrm{T}\\ &= \begin{bmatrix} 0 & \mathbf{x} & \mathrm{flip}(\overline{\mathbf{x}})\end{bmatrix}^\mathrm{T}\end{aligned} \tag{11}$$

and then define $\mathfrak{h}(\mathbf{x}) = \mathfrak{F}^{-1}(\mathfrak{s}(\mathbf{x}))$. Since $\mathfrak{s}(\mathbf{x})$ is conjugate symmetric, $\mathfrak{h}(\mathbf{x})$ is a real vector.

To verify if this conversion defines an equivalent scoring function for any triple $(r, s, o)$, let us now suppose complex embeddings $\mathbf{w}_r \in \mathbb{C}^n$ and $\mathbf{e}_s, \mathbf{e}_o \in \mathbb{C}^n$ are given. Since we regard real vectors $\mathfrak{h}(\mathbf{w}_r), \mathfrak{h}(\mathbf{e}_s), \mathfrak{h}(\mathbf{e}_o) \in \mathbb{R}^{2n+1}$ as the holographic embeddings of $r$, $s$ and $o$, respectively, the HolE score for the triple $(r, s, o)$ is given as

$$\begin{aligned}&f_{\mathrm{HolE}}(r, s, o)\\ &= \mathfrak{h}(\mathbf{w}_r) \cdot (\mathfrak{h}(\mathbf{e}_s) \star \mathfrak{h}(\mathbf{e}_o))\\ &= \frac{1}{n}\mathfrak{s}(\mathbf{w}_r) \cdot (\overline{\mathfrak{s}(\mathbf{e}_s)} \odot \mathfrak{s}(\mathbf{e}_o)) \qquad (\because \mathrm{Eq.\ (6)})\\ &= \frac{1}{n}\mathfrak{s}(\mathbf{w}_r) \cdot \begin{bmatrix} 0 & \overline{\mathbf{e}_s} \odot \mathbf{e}_o & \mathrm{flip}(\overline{\overline{\mathbf{e}_s} \odot \mathbf{e}_o})\end{bmatrix}^\mathrm{T} \qquad (\because \mathrm{Eq.\ (11)})\\ &= \frac{1}{n}\begin{bmatrix} 0 & \mathbf{w}_r & \mathrm{flip}(\overline{\mathbf{w}_r})\end{bmatrix}^\mathrm{T} \cdot \begin{bmatrix} 0 & \overline{\mathbf{e}_s} \odot \mathbf{e}_o & \mathrm{flip}(\overline{\overline{\mathbf{e}_s} \odot \mathbf{e}_o})\end{bmatrix}^\mathrm{T}\\ &= \frac{1}{n}\left(\mathbf{w}_r \cdot (\overline{\mathbf{e}_s} \odot \mathbf{e}_o) + \mathrm{flip}(\overline{\mathbf{w}_r}) \cdot \mathrm{flip}(\overline{\overline{\mathbf{e}_s} \odot \mathbf{e}_o})\right)\end{aligned}$$

$$= \frac{1}{n}\left(\mathbf{w}_r \cdot (\overline{\mathbf{e}_s} \odot \mathbf{e}_o) + \overline{\mathbf{w}_r} \cdot \overline{(\overline{\mathbf{e}_s} \odot \mathbf{e}_o)}\right)$$

$$= \frac{1}{n}\left(\mathbf{w}_r \cdot (\overline{\mathbf{e}_s} \odot \mathbf{e}_o) + \overline{\mathbf{w}_r \cdot (\overline{\mathbf{e}_s} \odot \mathbf{e}_o)}\right)$$

$$= \frac{2}{n}\operatorname{Re}\left(\mathbf{w}_r \cdot (\overline{\mathbf{e}_s} \odot \mathbf{e}_o)\right)$$

$$= \frac{2}{n}f_{\text{ComplEx}}(r, s, o),$$

which shows that $\mathfrak{h}(\cdot)$ (or $\mathfrak{s}(\cdot)$) gives the desired conversion from ComplEx to HolE.

## 6  Conclusion

In this paper, we have shown that the holographic embeddings (HolE) can be trained entirely in the frequency domain. If stochastic gradient descent is used for training, the conjugate symmetry of frequency vectors is preserved, which ensures the existence of the corresponding holographic embedding in the original real space (time domain). Also, this training method eliminates the need of FFT and inverse FFT, thereby reducing the computation time of the scoring function from $O(n\log n)$ to $O(n)$.

Moreover, we have established the equivalence of HolE and the complex embeddings (ComplEx): The spectral version of HolE is subsumed by ComplEx as a special case in which conjugate symmetry is imposed on the embeddings. Conversely, every set of complex embeddings can be converted to equivalent holographic embeddings.

Many systems for natural language processing, such as those for semantic parsing and question answering, benefit from access to information stored in knowledge graphs. We plan to further investigate the property of spectral HolE and ComplEx in these applications.

## References

Sören Auer, Christian Bizer, Georgi Kobilarov, Jens Lehmann, Richard Cyganiak, and Zachary Ives. DBpedia: A nucleus for a web of open data. In *The Semantic Web: Proceedings of the 6th International Semantic Web Conference and the 2nd Asian Semantic Web Conference (ISWC '07/ASWC '07)*, Lecture Notes in Computer Science 4825, pages 722–735. Springer, 2007.

Kurt Bollacker, Colin Evans, Praveen Paritosh, Tim Sturge, and Jamie Taylor. Freebase: A collaboratively created graph database for structuring human knowledge. In *Proceedings of the 2008 ACM SIGMOD International Conference on Management of Data (SIGMOD '08)*, pages 1247–1250, 2008.

Antoine Bordes, Jason Weston, Ronan Collobert, and Yoshua Bengio. Learning structured embeddings of knowledge bases. In *Proceedings of the 25th AAAI Conference on Artificial Intelligence (AAAI '11)*, pages 301–306, 2011.

Lise Getoor and Ben Taskar. *Introduction to Statistical Relational Learning*. Adaptive Computation and Machine Learning. The MIT Press, 2007.

Kelvin Guu, John Miller, and Percy Liang. Traversing knowledge graphs in vector space. In *Proceedings of the 2015 Conference on Empirical Methods in Natural Language Processing (EMNLP '15)*, pages 318–327, 2015.

Mohammad Al Hasan and Mohammed J. Zaki. A survey of link prediction in social networks. In Charu C. Aggarwal, editor, *Social Network Data Analytics*, chapter 9, pages 243–275. Springer, 2011.

Katsuhiko Hayashi and Masashi Shimbo. On the equivalence of holographic and complex embeddings for link prediction. In *Proceedings of the 55th Annual Meeting of the Association for Computational Linguistics (ACL '17): Short Papers*, pages 554–559, 2017. https://doi.org/10.18653/v1/P17-2088.

David Liben-Nowell and Jon Kleinberg. The link prediction problem for social networks. In *Proceedings of the 12nd Annual ACM International Conference on Information and Knowledge Management (CIKM '03)*, pages 556–559, 2003.

Maximilian Nickel, Kevin Murphy, Volker Tresp, and Evgeniy Gabrilovich. A review of relational machine learning for knowledge graphs. *Proceedings of the IEEE*, pages 1–18, 2015.

Maximilian Nickel, Lorenzo Rosasco, and Tomaso Poggio. Holographic embeddings of knowledge graphs. In *Proceedings of the 30th AAAI Conference on Artificial Intelligence (AAAI '16)*, pages 1955–1961, 2016.

Julius O. Smith, III. *Mathematics of the Discrete Fourier Transform (DFT): with Audio Applications*. W3K Publishing, 2nd edition, 2007.

Richard Socher, Danqi Chen, Christopher D. Manning, and Andrew Y. Ng. Reasoning with neural tensor networks for knowledge base completion. In *Advances in Neural Information Processing Systems 26 (NIPS '13)*, pages 926–934, 2013.

Fabian M. Suchanek, Gjergji Kasneci, and Gerhard Weikum. YAGO: A core of semantic knowledge unifying Wordnet and Wikipedia. In *Proceedings of the 16th International Conference on World Wide Web (WWW '07)*, pages 697–706, 2007.

Théo Trouillon, Christopher R. Dance, Éric Gaussier, and Guillaume Bouchard. Decomposing real square matrices via unitary diagonalization. arXiv.math eprint 1605.07103, arXiv, 2016a.

Théo Trouillon, Johannes Welbl, Sebastian Riedel, Éric Gaussier, and Guillaume Bouchard. Complex embeddings for simple link prediction. In *Proceedings of the 33rd International Conference on Machine Learning (ICML '16)*, pages 2071–2080, 2016b.

Bishan Yang, Wen tau Yih, Xiaodong He, Jianfeng Gao, and Li Deng. Embedding entities and relations for learning and inference in knowledge bases. In *Proceedings of the 3rd International Conference on Learning Representations (ICLR '15)*, 2015.